\documentclass[twocolumn]{article}
\usepackage{booktabs}

\usepackage{amssymb,amsmath,array}
\usepackage{tabularx}
\usepackage[T1]{fontenc}
\usepackage{natbib}
\usepackage{url}
\usepackage{geometry}
\usepackage{graphicx}
\graphicspath{{pics/},{.}}
\DeclareGraphicsExtensions{.pdf}

\usepackage{fancyhdr}
 
\usepackage{enumitem}

\newcommand{\comment}[1]{{}}
\def\bbbr{{\rm I\!R}} 
\newcommand{\R}{\bbbr}

\newcommand{\X}{\mathbf{X}}
\newcommand{\x}{\mathbf{x}}
\newcommand{\W}{\mathbf{W}}
\newcommand{\w}{\mathbf{w}}

\newcommand{\Y}{\mathbf{Y}}
\newcommand{\y}{\mathbf{y}}

\newcommand{\N}{\mathcal{N}}
\newcommand{\eps}{\boldsymbol{\epsilon}}
\renewcommand{\c}[1]{\mathcal{#1}}

\usepackage{times}
\long\def\@makecaption#1#2{\vspace{\abovecaptionskip}%
  \begingroup
  \footnotesize
  \textbf{#1.}\enskip{#2}\par
  \endgroup}

\usepackage[font=small,labelfont=bf]{caption}

\begin{document}

\title
{Sparse group factor analysis for biclustering of multiple data sources}
\author
{Kerstin Bunte\,$^{1,2}$, Eemeli Lepp\"aaho\,$^{1}$, Inka Saarinen\,$^{1}$, and Samuel Kaski\,$^1$\\
$^{1}$Helsinki Institute for Information Technology HIIT, Aalto University, Finland\\
$^{2}$Presently at School of Computer Science, University of Birmingham, Edgbaston, UK
}



\maketitle


\begin{abstract}
Motivation:
Modelling methods that find structure in data are necessary with the current large volumes of genomic data, 
and there have been various efforts to find subsets of genes exhibiting consistent patterns over subsets of treatments. 
These biclustering techniques have focused on one data source, often gene expression data. 
We present a Bayesian approach for joint biclustering of multiple data sources, 
extending a recent method Group Factor Analysis (GFA) to have a biclustering interpretation with additional sparsity assumptions. 
The resulting method enables data-driven detection of linear structure present in parts of the data sources.
Results:
Our simulation studies show that the proposed method reliably infers biclusters from heterogeneous data sources. 
We tested the method on data from the NCI-DREAM drug sensitivity prediction challenge,
resulting in an excellent prediction accuracy.  
Moreover, the predictions are based on several biclusters which provide insight into the data sources, 
in this case on gene expression, DNA methylation, protein abundance, exome sequence, functional connectivity fingerprints and drug sensitivity.
Availability: \url{http://research.cs.aalto.fi/pml/software/GFAsparse/}

\end{abstract}

\vspace*{-0.4cm}
\section{Introduction}

Numerous clustering approaches have advanced to extract knowledge from sets of e.g.\ gene expression experiments, 
when conditions of the samples are either not known or researchers are interested in dependencies within or across experiments.  
Conditions or treatments can affect the expression levels of certain genes only, and similarly, 
many genes are likely to be co-regulated under certain conditions only. 
For this purpose, biclustering techniques 
have been developed \citep{cheng00,hartigan72,lazzeroni02,morgan63}. 
Biclustering is traditionally defined as simultaneously clustering both rows and columns in a data matrix. 
Depending on the metric and the data, different approaches have emerged, aiming to cluster genes based on their expression levels being the same, 
differing by a constant, or being linearly dependent, with respect to different conditions \citep{Madeira:2004:BAB:1024308.1024313}. 
\cite{hochreiter2010fabia} introduced a generative approach called Factor Analysis for Bicluster Acquisition (FABIA), 
accounting for linear dependencies between gene expression and conditions. 
The biclusters are factors of the measurement matrix, and hence can be overlapping in both genes and conditions, 
whereas many approaches are limited to distinct clusters. 
Each bicluster can also include oppositely regulated genes (up- and down-regulated) across conditions. 
Similar approaches have been proposed by \cite{carvalho2008high} and \cite{gao2014differential}, 
with the latter one additionally focusing on the inference of gene co-expression networks. FABIA has also been extended to better suit genotype data \citep{Hochreiter2013}.

\cite{Waltman2010} proposed an algorithm for simultaneous biclustering of heterogeneous multiple species data collections. 
They investigate the identification of conserved co-regulated gene groups (modules) by comparing genome-wide datasets for closely related organisms 
and the evolution of gene regulatory networks. 
Most genes are unlikely to be co-regulated under every possible condition, and exclusive gene clusters cannot capture the complexity of transcriptional 
regulatory networks.
Their proposed approach aims to identify meaningful condition-dependent conserved modules, integrating data across the same genes present in multiple species.

Inferring bicluster structure jointly from multiple data sources is potentially more accurate than analysis of a single set, 
and the discovered relationships between the sources may offer new insights. 
In this paper, we extend a recent generative Bayesian modelling approach, group factor analysis (GFA) \citep{Virtanen12,suvitaival14,Klami14gfa}. 
GFA was developed for exploratory analysis of multiple data sources (views), resulting in an interpretable group-sparse factorization of the data collection. 
When the factors are additionally variable-wise sparse, as a result of introducing suitable priors, they are interpretable as biclusters of multiple
co-occurring data sources that need not share the same features (genes; as opposed to \cite{Waltman2010}). 
As a factor model GFA further shares the favourable properties of FABIA. 
We demonstrate its use in a multi-view drug sensitivity prediction task that the previous bicluster methods could not handle naturally: 
the approach shows superior prediction performance, 
and is able to infer meaningful structure present in subsets of the data.

\vspace*{-0.5cm}
\section{Methods}

\subsection{Factor analysis}
Factor Analysis for Bicluster Acquisition \citep{hochreiter2010fabia} assumes preprocessed and filtered gene expression data $\Y\in \R^{N\times D}$. 
Every row represents a sample and every column a gene. Therefore the value $y_{i,j}$ corresponds to the expression level of the $j$th gene in the $i$th sample. 
A bicluster is defined as a set of rows that are similar for a set of columns, and vice versa. 
The model for $K$ biclusters is
\begin{align}
\Y = \sum_{k=1}^K \x_{:,k} \w_{:,k}^\top + \eps \enspace,
\label{eq:FABIA_biclusters}
\end{align}
where each factor $k$ is defined by an outer product of the $k$th
columns of the factor matrix $\X\in\R^{N\times K}$ and the loading
matrix $\W\in\R^{D\times K}$, and $\eps\in\R^{N\times D}$ is normally
distributed noise: $\epsilon_{n,d} \sim \c{N}(0,\sigma_{d}^2)$. The factors
are the biclusters, with $|\w_{d,k}|$ indicating the (soft)
membership of gene $d$ in bicluster $k$, and $|\x_{n,k}|$ likewise for
sample $n$. 
Both the $\W$ and $\X$ are given sparse priors, more specifically component-wise
independent Laplace distributions:
\begin{align}
p(\W) &= \left(\frac{1}{\sqrt{2}}\right)^{DK} \prod_{k=1}^K \prod_{d=1}^D e^{-\sqrt{2}|\w_{d,k}|} \\
p(\X) &= \left(\frac{1}{\sqrt{2}}\right)^{NK} \prod_{k=1}^K \prod_{n=1}^N e^{-\sqrt{2}|\x_{n,k}|} \enspace .
\end{align}
The parameters are inferred with variational expectation maximization, 
see \citep{hochreiter2010fabia} for details.

\subsection{Sparse group factor analysis for biclustering}
\label{sec:GFA}
Group factor analysis has been proposed as an extension to factor
analysis for finding factors capturing joint variability between data
sets instead of individual variables \citep{Virtanen12,suvitaival14,Klami14gfa}.
It is designed to deal with several data sources
$\Y^{(1)}\in\R^{N\times D_1},\dots, \Y^{(M)}\in\R^{N\times D_M}$
(called views) of dimensionality $D_m$ with $N$ co-occurring
observations.  GFA models the $n$th sample of the $m$th data view as
\begin{equation}
\y^{(m)}_n \sim \N\left(\W^{(m)} \x_n, \tau^{-1}_m \mathbf{I}_{D_m}\right) \enspace ,
\label{eq:genModel}
\end{equation}
where $\tau_m$ is the noise precision of view $m$. The loading matrix
$\W^{(m)}$ is given a sparse prior that allows omitting any 
component $k$ from affecting drug data view $\Y^{(m)}$. This enables a
group-sparse factorization, where components may be (i) specific to a single
data view, (ii) shared between all the data views or (iii) shared
between any subset of the data views. As we are interested in finding
biclusters, we introduce a prior that is additionally variable-wise
sparse, that is, across the elements of the matrices $\W^{(m)}$ and
$\X$. This is done similarly to how 
\cite{khan2014identification} produced variable-wise sparsity, but now for both variables and samples to produce biclusters. 
Namely we use 
the following spike and slab priors
\footnote{\cite{suvitaival14} included sparsity for samples and mentioned the connection to biclustering, 
but the interpretation was not explored further}:
\begin{align}
x_{n,k} &\sim h_{n,k}^{(x)} N \left( 0, \left(\alpha_{k}^{(x)} \right)^{-1} \right) + \left( 1 - h_{n,k}^{(x)} \right) \delta_0 \label{eq:gfa1}\\
w_{d,k}^{(m)} &\sim h_{d,k}^{(m)} N \left( 0, \left(\alpha_{k}^{(m)} \right) ^{-1} \right) + \left( 1 - h_{d,k}^{(m)} \right) \delta_0 \\
h_{n,k}^{(:)} &\scriptstyle\sim \text{Bernoulli}(\pi_k^{(:)}) \quad
\pi_k^{(:)} \sim \text{Beta}(a^\pi, b^\pi) \quad
\alpha_{k}^{(:)} \sim \text{Gamma}(a^\alpha, b^\alpha)
\label{eq:gfa2}
\end{align}
where the binary $h^{(m)}_{d,k}$ determines whether the component
(bicluster) $k$ is active in the $d$th feature of $\Y^{(m)}$ (for all
non-zero $n$ in $h^{(x)}_{n,k}$), $\alpha^{(m)}_k$ determines the
scale of the component $k$ in view $m$ and $\pi^{(m)}_k$ the
probability of $h^{(m)}_{d,k}=1$. The prior is analogous for samples
(i.e. the rows of the data matrices) through $h_{n,k}^{(x)}$.
Effectively, the spike and slab prior will set weights that affect the data
(likelihood) only little to 0, 
which allows direct biclustering interpretations without a need for arbitrary thresholding afterwards.
The model is completed with a gamma prior for the noise
precision parameters $\tau_m$ and uninformative hyperpriors
($[a^\pi,b^\pi,a^\alpha,b^\alpha,a^\tau,b^\tau] = \mathbf{1}$). 

In this formulation, the data source information (feature grouping) is
used in three ways: (i) the noise precision ($\tau_m$) is the same for
all the features in a view, (ii) the binary vector $h^{(m)}_{:,k}$ has
a common probability ($\pi^{(m)}_k$) of being active and (iii) the
scale of a component ($\alpha^{(m)}_k$) is shared within a view. With
an uninformative Gamma prior, often called Automatic Relevance
Determination prior, this third property implements the group
sparsity.  The second property implies that a specific feature $d$ (in
view $m$) is more likely to be active in a bicluster, if many of the
features in view $m$ belong to the said bicluster, and vice versa. 
This allows explaining variance that is not present in all the data sources, but dense in some of them, more robustly. 
Given data with significant (source specific) structured variation respective components can help to detect biclusters more accurately. 

The formulation above assumes that all the data views have co-occurring samples. 
We also extend GFA for joint modelling of data sets that are paired in two modes (see Fig.~\ref{fig:toy2way}), 
i.e. $\{\Y^{(1,1)},\dots, \Y^{(M_1,1)},$ $\Y^{(1,2)}, \dots \Y^{(M_2,2)}\}$, 
where $\Y^{(m,2)} \in \R^{D_1,N_2}$ is paired with the features of $\Y^{(1,1)}$. 
Both the modes will have a set of components identical to the ones presented above with one exception, and hence 
we will not repeat the details of the priors here. 
The exception is that the view paired in both the modes is generated from the components of both the modes, as
\begin{equation}
y^{(1,1)}_{i,j} \sim \N\left(\w^{(1,1)}_j \x^{(1)}_i + \w^{(1,2)}_i \x^{(2)}_j, \tau^{-1}_{1,1} \right) \enspace .
\end{equation}
As the priors remain conjugate, the model can be inferred using Gibbs sampling, resulting in linear complexity in both $N$ and $D$, 
but cubic w.r.t. the number of components $K$. 
The parameters shown in this paper, and used in predictive tasks, will be the posterior means. 
In the following sections we will test FABIA and GFA in 
several simulation studies and a drug sensitivity analysis, incorporating genetic data available from the DREAM project.

\vspace*{-0.3cm}
\section{Simulation study}

\begin{figure}[!t]
\begin{picture}(225,150)
\put(5,10){\includegraphics[width=0.47\textwidth]{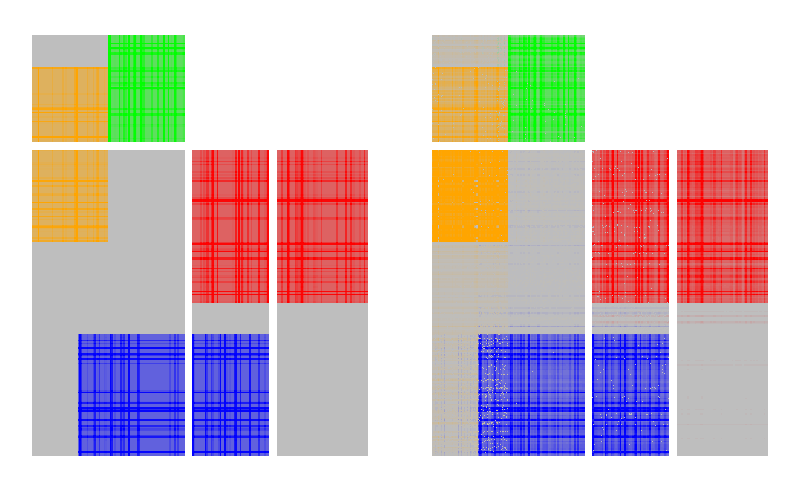}}
\put(0,60){$N_1$} \put(0,130){$N_2$} \put(30,15){$D_1$} \put(67,15){$D_2$} \put(95,15){$D_3$}
\put(25,60){$\Y^{(1,1)}$} \put(25,130){$\Y^{(1,2)}$} \put(62,60){$\Y^{(2,1)}$} \put(90,60){$\Y^{(3,1)}$}
\put(143,60){$\Y^{(1,1)}$} \put(143,130){$\Y^{(1,2)}$} \put(180,60){$\Y^{(2,1)}$} \put(208,60){$\Y^{(3,1)}$}
\end{picture}\vspace*{-0.5cm}
\caption{
{\bf Left}: 4 non-overlapping biclusters (coloured blocks) used in 
the multi-view data (gray area). 
{\bf Right:} The biclusters inferred by GFA.}
\label{fig:toy2way}
\end{figure}
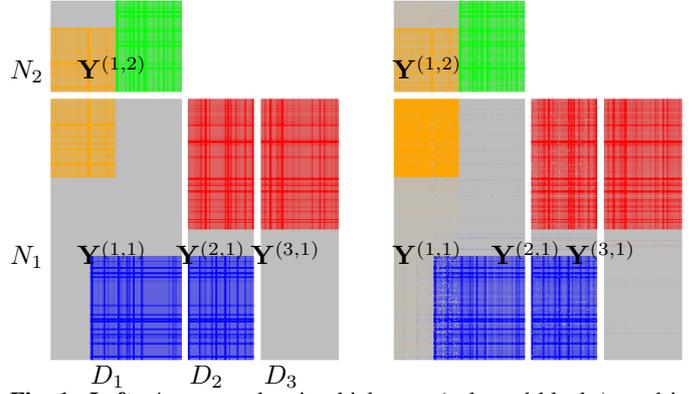
We show an illustrative example of bicluster inference, generating
a collection of data sets, $\{\Y^{(1,1)},\Y^{(2,1)},\Y^{(3,1)},\Y^{(1,2)}\}$, with 200 samples and dimensions
$(100,50,60)$ for $\Y^{(:,1)}$, and 100 samples and dimension $(70)$ for $\Y^{(1,2)}$. 
The data collection was generated with four biclusters and additional noise with variance 1. 
The non-zero parts of $\x$ and $\w$ for the biclusters were drawn from $\N(0,1)$, 
but truncated between absolute values 1 and 2 for illustrative purposes. 
The data structure is shown in Fig.~\ref{fig:toy2way} (left); for clarity the biclusters are non-overlapping blocks. 
We inferred the component structure of these data using GFA; the posterior mean of the biclusters is visualized in Fig.~\ref{fig:toy2way} (right). 
GFA can clearly infer this kind of component structure very accurately.

GFA has been designed for joint modelling of multiple data sets. 
However, when the data consist of one set only, FABIA and GFA are essentially the same model; 
in the current implementations there is the technical difference that FABIA has a continuous valued sparsity prior
for $\X$ and $\W$, whereas GFA implements a discrete choice with the spike-and-slap. 
We first investigate the effect of this technical difference by comparing GFA
with FABIA on single-view data (FABIA1), and then investigate how much multi-view data helps, 
by comparing GFA against FABIA for which data are concatenated into a single matrix (FABIA2).

For the simulation studies, we construct data from the generative
model Eq.~\eqref{eq:genModel}, with matrices $\X$ and $\W$ generated
such that each element is either zero or sampled randomly from the
normal distribution $\mathcal{N}(0,1)$ to build the bicluster(s).  The
resulting data matrices $\Y$ are given to the methods, which then
return the bicluster estimates
$\x^\prime_{:,k}\w^{\prime\top}_{:,k}$. They are compared to the true
biclusters to analyse the models' performance.  FABIA is run with the
correct number of biclusters $K$, and the results are reported for a
range of thresholds.  GFA learns the cluster number by
driving unnecessary ones to zero, and we used a component number 5 above
the correct $K$. The final biclustering is based on 101 posterior
samples (2000 burn-in samples, 20 thinning): if the majority of 
$(\x_{:,k}\w_{:,k}^{(m)\top})_{ij}$ are non-zero in the posterior
samples, then $\Y^{(m)}_{ij}$ is assigned to the $k$th bicluster,
otherwise not.  All the simulation studies are repeated 10 times and
we report the average $F_1$ score for detecting the true bicluster
structure:
\begin{align}
F_1 = \frac{2 TP}{2TP + FN + FP} \,
\label{eq:F}
\end{align}
where $TP$, $FN$ and $FP$ denote the number \emph{true positives}, \emph{false negatives} and \emph{false positives}, respectively, summed over all the elements 
of the data matrix. To be able to evaluate the effects of different sparsity priors and inference techniques, we also compare to FA with similar priors as GFA, using the full 
concatenated data, and inferred with Gibbs sampling. This is done be changing Eq. \eqref{eq:gfa1} to Eq. \eqref{eq:gfa2} to allow only single $\alpha^{(1)}_k$ and $\pi^{(1)}_k$ 
across the data features, and by giving each feature independent noise precision $\tau_d$.

\begin{figure}[t]
\includegraphics[width=0.98\columnwidth]{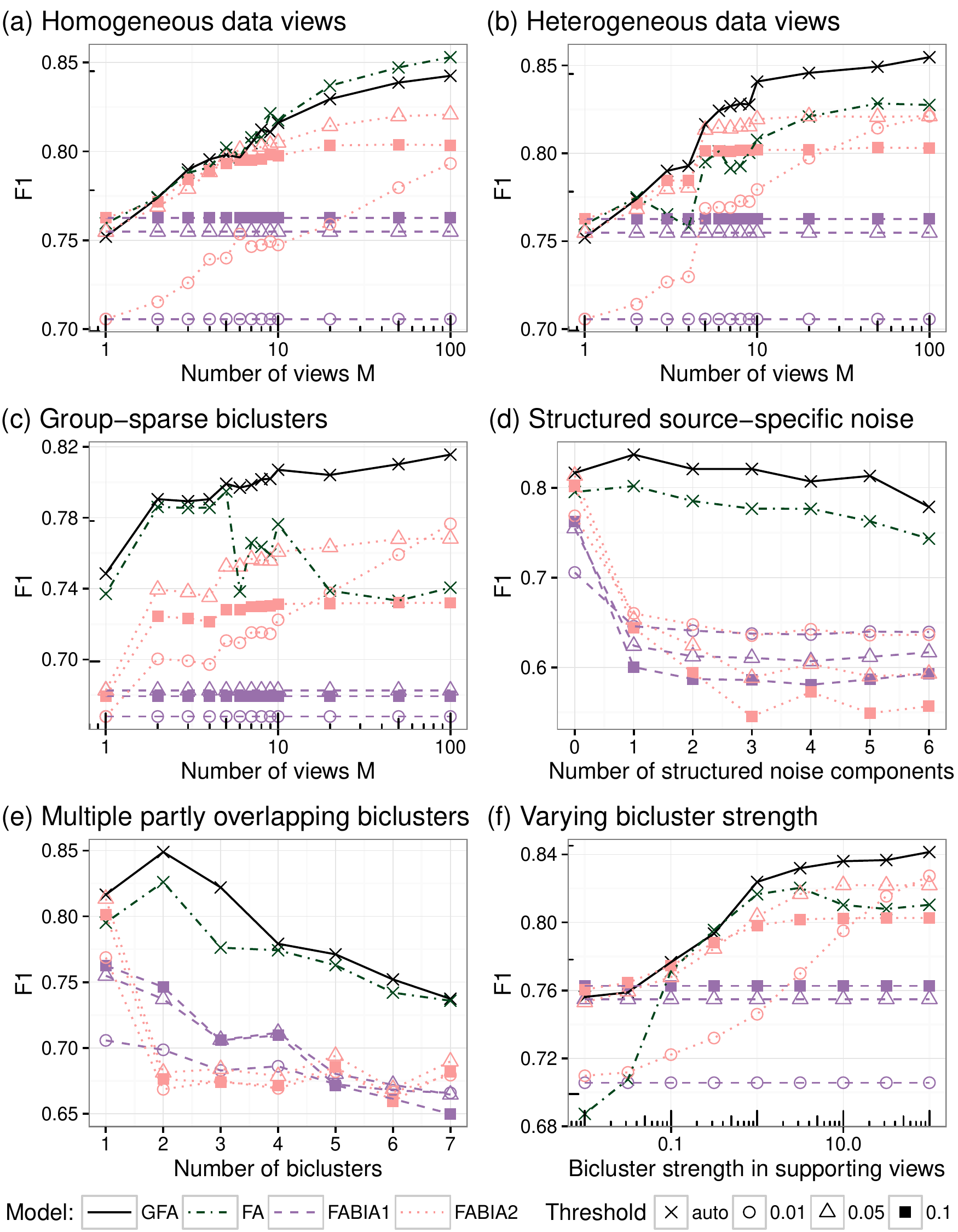}
\caption{Simulated experiments comparing the abilities of GFA, FA and FABIA to detect data-generating biclustering. 
(a) and (b) report the $F1$ scores over a varying number of data views ($M$) present, for homogeneous and heterogeneous data collections, respectively.
The biclusters are further assumed group-sparse in (c). In (d) the problem is made more challenging by adding structured noise on top of the signal,
whereas the number of biclusters is varied in (e). 
In (f), the strength of the bicluster is varied in the supporting views. FABIA1 uses only the data matrix of interest, 
whereas FABIA2 and FA have side information concatenated with it;
both FABIAs are reported for thresholds $(0.01, 0.05, 0.10)$ determining the biclusters.
}
\label{fig:ToyData}
\end{figure}

By default we use $M=5$ data views, $N=50$ samples, $D_m=100$ features per data view and one bicluster active in $70\%$ of the samples and the features. 
Bicluster and noise variance in the view of interest are set to 1.
To make the data views heterogeneous, the other 4 data views are given bicluster and noise variances of $((0.2,0.2),(5,5),(0.2,5),(5,0.2))$. 
We report the mean performance of the methods in Fig.~\ref{fig:ToyData}, in six different experimental settings:
\begin{enumerate}[leftmargin=8pt,label=(\alph*),noitemsep,topsep=0ex,labelsep=2pt]
\item All data views are set to be homogeneous having variance of 1 in the bicluster and the noise residual, 
and the number of data views is varied. Due to the homogeneity of the data views, GFA has no advantage over FA. 
FA outperforms FABIA on high-dimensional large number of views.
\item Similar to (a), but with heterogeneous data with respect to the bicluster and noise residual strength. 
The multi-view approach of GFA is superior, 
while FA and FABIA have similar performance.
\item Similar to (b), but with sparse biclusters 
w.r.t.\ the samples and group sparse w.r.t.\ the features (present in all but every 3rd view). 
  This matches GFA's assumptions leading to superior results.
\item Additional view-specific noise components (0 to 6) were added on
  top of the bicluster signal (as $\x_\text{noise}\w_\text{noise}^\top$, each vector element sampled from $\N(0,1)$).
  The GFA-type of priors and inference are clearly
  more robust against the structured noise.
\item We evaluated the accuracy of detecting $1$ to $7$ partly overlapping biclusters. GFA outperforms FA, 
whereas FABIA does not seem very robust when the number of biclusters increases; using the additional data can even decrease its performance.
\item The strength of the biclusters is varied for the additional views
  (precision $\alpha$ ranging from $10^{-2}$ to $10^2$). GFA is more
  accurate when the additional views have strong biclusters, and has a
  similar performance with FABIA when weaker. FA suffers compared to FABIA
  when the additional views get less relevant.
\end{enumerate}

Across all the studies discussed above, GFA was able to detect the correct number of biclusters (and additional noise components) exactly
in 90.4$\%$ of the runs, and overestimated it by 1, 2 or 3 in $9\%$, $0.4\%$ and $0.1\%$ of the runs, respectively. 
The extra components were modelling artificial structure detected in the residual noise and did not resemble the bicluster structure. 
To confirm that they did not give an advantage in the comparison, we also ran FABIA with 5 extra components, 
resulting in consistently worse performance compared to the reported runs, where FABIA was given the true component amount.
The standard deviations of the mean $F1$ scores, averaged over the 10 independent repetitions, ranged from 0.003 to 0.01.
Inferring a single model took on average 22s, 45s, 0.04s, and 0.5s for GFA, FA, FABIA1 and FABIA2, respectively, demonstrating the efficiency of the EM-algorithm in FABIA.

Our FA implementation is generally, but not consistently, more robust than FABIA in bicluster detection with additional data sources.
The advantages of the multi-view setup of GFA (vs concatenation in FA and FABIA2) are most significant
when (i) there are plenty of heterogeneous data views, (ii) the biclusters are group-sparse, or (iii) the data views are highly heterogeneous. 
These conditions are realistic in real-life applications.

\vspace*{-0.5cm}
\section{Drug response study}

The NCI-DREAM drug sensitivity prediction challenge \citep{costello2014community} provided 
publicly available data consisting of gene expression (GE), RNA, DNA methylation (MET), copy number variation (CNV),
protein abundance (RPPA) and exome sequence (EX) measurements for 53 human breast cancer cell lines. 
In the challenge, expression data were based on Affymetrix Genome-Wide Human SNP6.0 Array and Affymetrix GeneChip Human Gene 1.0 ST microarrays.
RNA sequencing libraries were prepared using the TruSeq RNA Sample Preparation Kit and whole transcriptome shotgun sequencing was performed. 
The Mutation status was obtained from exome-capture sequencing and 
GenomeStudio Methylation Module v1.0 was used to express the methylation for each genome-wide detected CpG locus resulting in values between 
0 (completely unmethylated) to 1 (completely methylated) proportional to the degree of methylation at any particular locus. 
More details on the preparation of the genomic data for the challenge are provided by \cite{costello2014community}.
Each cell line was exposed to 31 therapeutic compounds and the dose-response values of growth inhibition were collected. 
The drug response data was revealed only for 35 of the cell lines, and the challenge was to predict the
response of the remaining 18 cell lines, ranking them from the most sensitive to the most resistant.  

As the drug response prediction problem is extremely challenging, we performed the following  
steps, learning from \cite{costello2014community},
to increase the signal-to-noise ratio: We reduced the dimensionality to the 500 genes with the highest average variance over the data views, 
including the overlapping set of 14 genes appearing in the RPPA data set.
Furthermore, the most predictive data sources for further analysis were chosen by 7-fold cross validation on the 35 training samples with known drug response values.
To compare the sources the root mean squared error (RMSE), as well as Pearson and Spearman correlations of the predicted drug responses, were computed 
averaged over 10 repetitions of the experiments, each with different random splits. We inferred the GFA model for this multi-view data by ignoring the missing drug response data in 
the likelihood,
after which the missing values can be predicted from $\X\W^{(m)}$, where $\X$ for the missing cell lines is inferred based on the other data sources only.
The performance of GFA trained with different combinations of data views is shown in Table \ref{tab:viewExps}.

\begin{table}
\caption{Averaged 7-fold cross validation results for GFA on the training set of the DREAM7 drug sensitivity prediction challenge to
  identify the views showing best prediction performance for further analysis.}
\label{tab:viewExps}
\begin{tabular*}{0.98\linewidth}{@{\extracolsep{\fill}} lrrr}
Views used & RMSE & Pearson & Spearman \\\toprule
All & 1.9 & 0.031 & 0.079 \\
\hline
GE, MET, CNV, RNA, RPPA & 2.3 & 0.016 & 0.088 \\
GE, CNV, RNA, RPPA, EX & 2.0 &  0.031 & 0.078 \\
GE, MET, CNV, RPPA, EX & 1.5 & 0.040 & 0.085 \\
GE, MET, CNV, RNA, EX & 1.8 & 0.012 & 0.078 \\
MET, CNV, RNA, RPPA, EX & 1.6 & 0.018 & 0.058 \\
GE, MET, RNA, RPPA, EX & 1.9 & 0.040 & 0.089 \\
\hline
\uppercase{ge, met, cnv, rppa} & 1.8 & 0.028 & 0.071 \\ 
\uppercase{ge, cnv, rppa, ex} & 1.8 & 0.018 & 0.074 \\ 
\uppercase{ge, met, cnv, ex} & 1.5 & 0.024 & \textbf{0.090} \\ 
\uppercase{met, cnv, rppa, ex} & 2.1 & 0.020 & 0.061 \\
\textbf{\uppercase{ge, met, rppa, ex}} & \textbf{1.4} & \textbf{0.046} & 0.087 \\
\hline
\uppercase{ge, met, rppa} & 1.9 & 0.024 & 0.072 \\
\uppercase{ge, rppa, ex} & 1.6 & 0.016 & 0.059 \\
\uppercase{ge, met, ex} & 1.5 & 0.042 & 0.084 \\
\uppercase{met, rppa, ex} & 1.8 & 0.011 & 0.075 \\\bottomrule\vspace*{-0.8cm}
\end{tabular*}
\end{table}

The most promising views finally chosen for the bicluster analysis were gene expression, methylation, exome sequence and RPPA measurements, 
leaving out the copy number variation and RNA.  
Finally, we ran GFA for the full data (handling the test drug responses as missing values) and reconstructed the 
missing data averaged over the posterior samples of 50 sampling chains. 
We gave the model a mildly informative prior assuming signal-to-noise ratio of 0.5. 
All the sampler chains were initialized with $K=60$, allowing data-driven inference of model complexity 
(resulting in 48 to 56 components). 
A total of 100 sampled parameters were stored for each chain (every 20th sample stored after
10000 burn-in iterations), resulting in an average runtime of 84 minutes per chain. 
The performance was quantified using the same score as in the challenge, that is, the weighted averaged probabilistic concordance index.  
We achieved a score of 0.592, which would have been placed the first in the challenge (winner model reaching 0.583),
indicating excellent prediction performance of GFA on this data, possibly stemming from the biclustering nature of the model. Furthermore, we ran
GFA utilizing data sources paired in two modes in a similar way with additional
functional connectivity fingerprints describing the drugs
(FCFP; calculated with PaDEL-Descriptor,~\cite{yap2011padel}), allowing joint modelling of biological and chemical effects in the measured data. 
The additional chemical view resulted in a slight increase in the target score, to 0.599.
The structure of the latter model is interpreted in the following sections, motivated by the excellent predictive performance.

\vspace*{-0.3cm}
\subsection{Robust components}

For interpretation purposes we next sought representative point solutions to describe the posterior distributions.
Due to the extremely challenging nature of the problem 
the total variance explained by individual components is small.
To minimize the risk of analysing patterns occurred by chance, we searched for components that occur consistently across the different sampling chains, 
making the assumption (which was checked manually) that component indices are reasonably stable within a chain, 
but can naturally be arbitrarily permuted between chains.
To find the matches between chains we averaged the components over the posterior samples within their chain, 
and compared using cosine similarity. 
If the similarity of the best match exceeded the threshold 0.80, we considered the components to be the same. 
Furthermore, we chose to further study components found in at least half of all chains, deemed robust in this procedure. 
Out of the average 52.6 components inferred by the sampling chains, 25 were on average chosen to the set of robust components. 
Ideally we would infer the model parameters with a single well-mixing sampling chain, but as the posterior is multimodal 
(and we do not want to constrain it artificially) the inference problem is extremely challenging, and 
we resort to the described computational simplification.
 
We observed that some of the components are very sparse, only containing one or two cell lines and hence most probably explaining
outliers in the data. 
Therefore, we will focus the interpretations on the more dense biclusters only; 3 out of the total 27 biclusters
found predicted 1.26\%, 0.06\% and 0.11\% of the total variance in the test data (2.89\%, 0.3\% and 0.98\% in the set of active drugs).
There was 1 additional bicluster shared with the drug descriptors, but it had no significant effect to the test data. 
With the drug sensitivity prediction of these four components only, we received a target score of 0.591.

\begin{figure*} 
\centering
\includegraphics[width=0.95\linewidth]{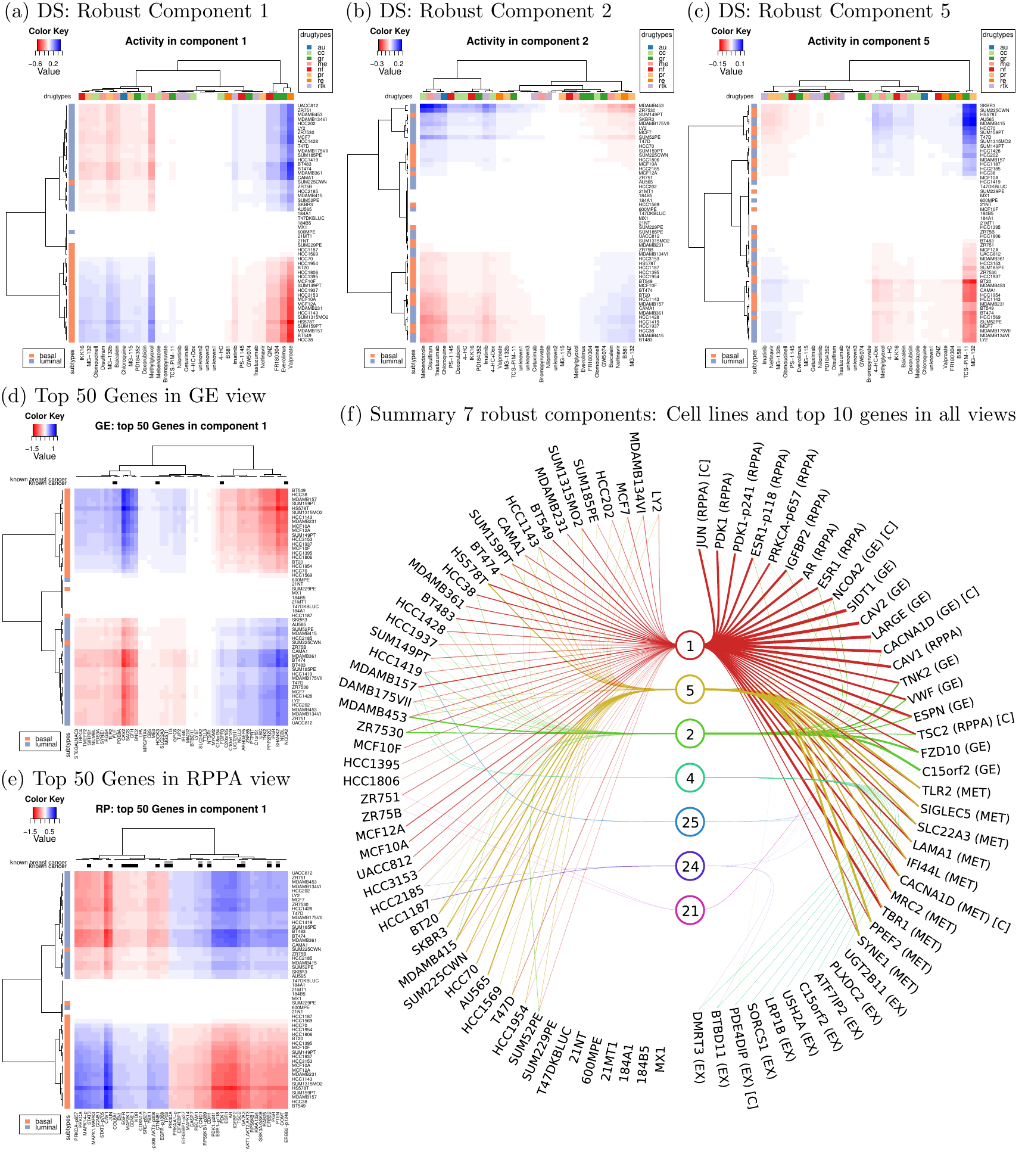}
\caption{Bicluster activity patterns of robust components. (a)-(c) show the intensity values of cells (in rows) and drugs (in columns) in the drug sensitivity (DS) view of 3 
different components. Component one mainly distinguishes basal and luminal cell lines, while (d) and (e) show the corresponding bicluster activity pattern of the 50 genes with 
highest mean absolute values in the RPPA and GE views, marking known (breast) cancer genes by a (gray) black square. (f) shows bicluster participation of all cells (left) in 7 
selected robust components (middle) with respect to their mean absolute intensity values (represented in the thickness of the lines) for the top 10 genes in each of the views (GE, 
MET, RPPA and EX). Known cancer genes are marked by [C].}
\label{fig:DreamFindings}
\vspace*{-0.4cm}
\end{figure*}

\vspace*{-0.3cm}
\subsection{Interpretations of the biclusters}

For interpretation purposes, we collected the descriptions of the drugs and cell lines used in the challenge \citep{costello2014community}. 
Some groups of drugs can be identified, which we will abbreviate as: autophagy (au), cell cycle (cc), metabolism (me), 
regulation (re) and signalling growth (gr) drugs, as well as nuclear factor (nf), protease (pr) and receptor tyrosine kinase (rtk) inhibitors.
Furthermore, most of the cell lines represent a subtype of cancer which can be categorized as basal or luminal.

The bicluster structure of the activity patterns for the first component in the drug sensitivity (DS) view, consisting of
measurements of sensitivity of cell lines to drugs, is depicted in Fig. \ref{fig:DreamFindings}a. 
Component one distinguishes basal and luminal cell types, without that information being used in the training. 
The response for all 5 cell cycle and all 4 metabolism drugs is positive or above average for most of the basal cell lines,
whereas luminal cells show negative activation. Luminal cells respond strongly to regulation drugs, 
where the response of basal cells is negative. 
Component 2 shows high activity patterns for proteasome and cell cycle drugs as depicted in Fig. \ref{fig:DreamFindings}b. 
The other components have relatively small biclusters with only a few active cells and drugs. 
Component 3, for example, shows cells that are (un)responsive to rtk inhibitors and otherwise mixed groups of cells
and drugs (see Fig.~\ref{fig:DreamFindings}c). 
The remaining robust component, in the second mode, was associated with most of the drugs and drug descriptors, 
and weakly with approximately half of the cell lines (strongly with T47DKBLUC). 

\begin{figure}
\centering
\includegraphics[width=0.96\linewidth]{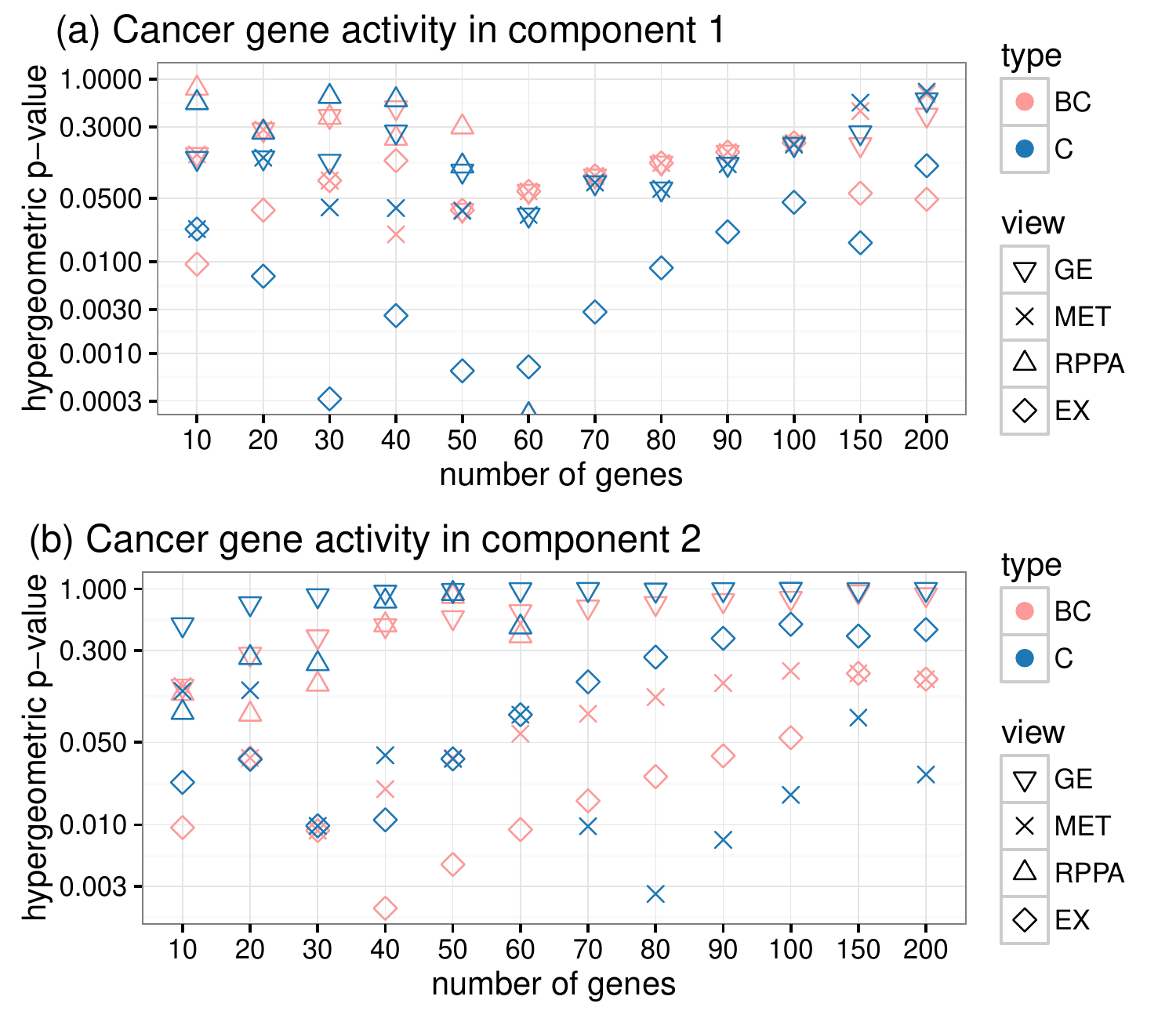}
\caption{Hypergeometric test of the activity of known breast cancer (BC) and all cancer (C) genes in the two most predictive components and all views. 
Low P-values indicate a high number of cancer genes in the top $n$ active genes in comparison with randomly picked sets. 
See the text for details.
}
\label{fig:allCancer_pValues}
\end{figure}

Due to the large number of genes in the other views, we show summaries of enrichment of known cancer genes in the components. 
We performed hypergeometric tests comparing a varying number of the most active genes (i.e. genes with the highest mean absolute values in $\W$ corresponding to RPPA and GE views) 
and random sets of equal size, 
for the occurrence of known cancer genes (extracted from \cite{Stephens2012}) in the most predictive components and all views. 
Low $p$-values indicate that the approach is able to detect a significant amount of known (breast) cancer genes in the top 
active genes of the components when compared to random subsets of genes in the views.
Fig. \ref{fig:allCancer_pValues} shows the results of the hypergeometric test for two robust components. 
For component 1 (Fig. \ref{fig:allCancer_pValues}a) we observe highly significant cancer activity already in the 10 most 
active genes in every view, except in the RPPA data.  
In the Exome sequence data we observe significant cancer gene activity independently of the size of the subset. 
For component 2 (Fig. \ref{fig:allCancer_pValues}b) we observe less significant activity of cancer genes in the gene 
expression and exome data than in component 1. 
However, we find high cancer gene activity in the methylation view and very high activity of breast cancer genes in exome sequencing data.

Besides the statistical tests we report the results on the level of individual genes.  
We condensed the analysis showing the 50 most active genes (in terms of their absolute values) in component 1 in the gene expression
(Fig. \ref{fig:DreamFindings}d) and the RPPA view in Fig. \ref{fig:DreamFindings}e. 
Component 1 contains proportional and anti-proportional co-regulated genes as indicated by the intensity of the biclusters depicted. 
The genes which are known as cancer or breast cancer genes are marked by a black and gray squares, respectively. 
Fig. \ref{fig:DreamFindings}f summarizes the mean participation in biclusters of the cells in seven different components on the top 10 active genes in each of the views. 
The left side contains the list of all cells sorted by their mean absolute values in the seven most active components, which are depicted in the middle row. 
The right side contains the list of top genes clustered by their mean activity in these components, accompanied by a
shortcut for the view they were taken from and [C] in case they are known as cancer gene in the literature.
Component one (coloured red) contains the biggest biclusters with comparably high mean absolute values depicted by thicker connecting lines 
in lots of cells and genes throughout the different views except the exome sequence data. 
Even only selecting 10 most active genes from each view delivers at least one known cancer gene. 
Although components overlap they also depict relationships of different cells in different views.

Furthermore, we performed a Gene Ontology (GO) enrichment analysis on the most active gene sets in the components and gene related views.
In GO the genes or gene products are hierarchically classified and grouped into three categories: \emph{molecular function} (mf)
describing the molecular activity of a gene, \emph{biological process}
(bp) denoting the larger cellular role and \emph{cellular component}
(cc) depicting where the function is executed in the cell. 
The enrichment analysis was performed directly in the GO website (\url{http://geneontology.org/}), which connects the 
PANTHER \citep{Mi2013} classification system with GO annotations.
From each of the gene-related views GE, MET, RPPA and EX we selected a list of the 50 most active genes from the dense robust components. 
For each of such gene sets we calculated the enrichment for all categories. 
The result table contains a list of shared GO terms for each gene set 
together with information about the background and sample frequency,
fold enrichment and the p-value determined by a binomial statistic.  
A p-value close to zero indicates the significance of the GO term associated with the provided group of genes.

\begin{figure}
\centering
\includegraphics[width=0.95\linewidth]{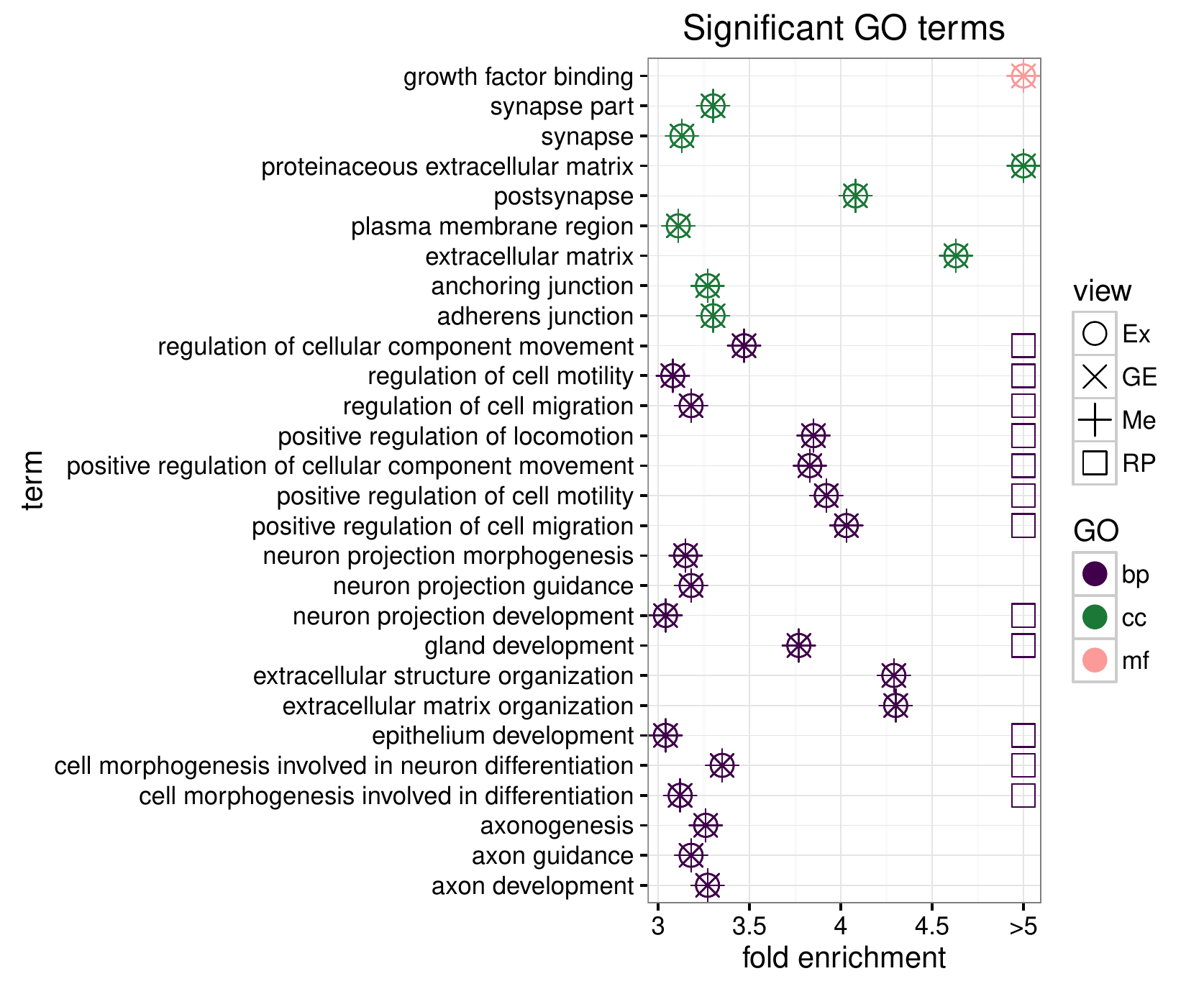}
\caption{Enrichment of the most significant GO terms, which occur more than 3 times in all the gene related views and the three dense robust components.}
\label{fig:significantGOterms}
\end{figure}

More than one thousand shared GO terms are returned for the most active gene sets. 
We condensed the results showing only the most repeating and most significant ones by 
using a threshold for the p-value and showing only GO terms below $10^{-6}$, which
appear throughout the views and components more than 3 times.
Fig. \ref{fig:significantGOterms} shows the reduced list of significant GO terms with fold enrichment value bigger than 3. 
This value indicates the magnitude of fold enrichment for the observed set of genes over the expected, 
thus with values bigger than one the category implying over-representation. 
For biological process we found most of the repeating significant GO terms in all of the views in nearly all cases.
These GO terms are related to cell motility and its regulations. 

\section{Discussion}

We presented sparse group factor analysis as a way of inferring biclusters from heterogeneous multi-source data.  
The method is able to detect predictive and interpretable structure present in any subset of the data sources, and sparse within the sources. 
It proved to be robust in this task, as witnessed by the simulation studies and the
outstanding performance in the NCI-DREAM drug sensitivity prediction challenge. 
The biclusters of the joint data identified cancer cell subtypes, 
grouped drugs by their functional mechanisms, and associated known cancer genes with the drug sensitivity data, all in a data-driven fashion. 
The shown approach is suitable for exploratory analysis of multiple data sources, 
giving condensed and interpretable information with respect to the data collection.\\
In this paper we focused on formulating a model that implements the novel multi-data-source biclustering, 
and on evaluating the accuracy of the results. 
Two important questions we did not yet fully address are: 
(i) Could some of the alternative ways of implementing sparsity, 
substituting the spike-and-slabs of this paper, result in computationally more efficient and still as accurate solutions. 
(ii) Computational speed. 
The EM point estimates the single-data-source FABIA algorithm uses would naturally be faster for multiple data sources as well, 
but the ability to handle uncertainty due to highly noisy and high-dimensional small sample-size data would suffer. 
Variational approximations would be attractive as they would also help avoid the matchings between the different sampling chains, 
but deriving variants of the algorithm would be more difficult and variational approximations are known to produce a biased estimate of the uncertainty of the solutions. 
For large data parallelised sampling solutions would be particularly attractive ways of speeding up computation.

\vspace*{-0.3cm}
\subsection*{Acknowledgement}
We thank the Academy of Finland (Finnish Centre of Excellence in Computational Inference Research COIN) for funding.

\vspace*{-0.5cm}
\bibliographystyle{natbib}
\bibliography{document}

\end{document}